\documentclass[conference]{IEEEtran}
\pdfoutput=1

\IEEEoverridecommandlockouts

\usepackage{cite}
\usepackage{amsmath,amssymb,amsfonts}
\usepackage{algorithmic}
\usepackage{graphicx}
\usepackage{textcomp}
\usepackage{xcolor}
\usepackage{subcaption}
\usepackage{comment}
\usepackage{float}

\usepackage{multirow}

\def\BibTeX{{\rm B\kern-.05em{\sc i\kern-.025em b}\kern-.08em
    T\kern-.1667em\lower.7ex\hbox{E}\kern-.125emX}}
    
\begin{document}

\title{FCN+RL: A Fully Convolutional Network followed by Refinement Layers to Offline Handwritten Signature Segmentation\\
}


\author{\IEEEauthorblockN{
Celso A. M. Lopes Junior\,$^{1}$,
Matheus Henrique M. da Silva\,$^{1}$,
Byron Leite Dantas Bezerra\,$^{1}$,\\
Bruno Jos\'e Torres Fernandes\,$^{1}$ and
Donato Impedovo\,$^{2}$
}
\IEEEauthorblockA{\,
$^{1}$Escola Polit\'ecnica de Pernambuco, Universidade de Pernambuco, Recife, Brasil\\
$^{2}$University of Bari, Bari, Italy\\
Emails: \{camlj, mhms\}@ecomp.poli.br, byron.leite@upe.br,
bjtf@ecomp.poli.br,
donato.impedovo@uniba.it}
}

\maketitle

\begin{abstract}
Although secular, handwritten signature is one of the most reliable biometric method used by most countries.
In the last ten years, the application of technology for verification of handwritten signatures has evolved strongly, including forensic aspects. Some factors, such as the complexity of the background and the small size of the region of interest - signature pixels - increase the difficulty of the targeting task. Other factors that make it challenging are the various variations present in handwritten signatures such as location, type of ink, color and type of pen, and the type of stroke. In this work, we propose an approach to locate and extract the pixels of handwritten signatures on identification documents, without any prior information on the location of the signatures. The technique used is based on a fully convolutional encoder-decoder network combined with a block of refinement layers for the alpha channel of the predicted image. The experimental results demonstrate that the technique outputs a clean signature with higher fidelity in the lines than the traditional approaches and preservation of the pertinent characteristics to the signer's spelling. To evaluate the quality of our proposal, we use the following image similarity metrics: SSIM, SIFT, and Dice Coefficient. The qualitative and quantitative results show a significant improvement in comparison with the baseline system.
\end{abstract}

\section{Introduction}
\label{intro}

Handwritten signature is a biometric authentication method widely used for personal documents and legal contract validations. Besides, experts in forensic analysis examine handwritten signatures to certify the authenticity of the writing and reveal possible fraud, which in some cases can mean high-value financial losses.
 
One possible approach to signature authentication is through human operators who must compare the signature present in the document with the signature of the original subscriber. However, this approach can be expensive and time-consuming, given the amount of data accumulated in institutions that use a handwritten signature as a means of identification and authentication \cite{melo2018fully}.

Several approaches have been developed in the field of machine learning and statistical methods to perform the signature detection and verification tasks automatically. Among these approaches, we can mention techniques based on Artificial Neural Networks (ANN) \cite{pansare2012handwritten}, Hidden Markov Models (HMM) \cite{coetzer2004offline}, Support Vector Machines (SVM) \cite{matsuda2016effective}, and Fuzzy Logic \cite{anikin2016handwritten}. Among the neural network techniques, it is important to mention Faster Region-based Convolutional Neural Networks (RCNN) \cite{ren2015faster} and YOLOv2 \cite{redmon2017yolo9000}. Both models were adapted for logos and signature localization in noisy documents \cite{sharma2018signature}. 

Many of the techniques that address signature verification use public databases \cite{guerbai2015effective}\cite{hafemann2017learning}\cite{soleimani2016deep}\cite{diaz2016generation}. However, such bases as GPDS \cite{vargas2007off}, and MCYT \cite{ortega2003mcyt} present images with a light background and dark signatures. This  characteristic does not present an environment of great complexity for segmenting the signature pixels. Besides, due to the insertion of mobile devices and their growing popularity, several commercial and banking applications, for example, use images captured by smartphones for transactions, payments, account opening, and copies of documents \cite{diaz2019perspective}\cite{mello2012}. 

On the other hand, document images captured by smartphone cameras are usually presented with distortions and background noise. Therefore, treating these images in such a way that only handwritten signatures can be extracted for analysis of their characteristics becomes a challenging task in image processing. These images do not always present the desired features or the expected quality, negatively influencing the process of recognition and classification of these handwritten signatures.

The situation may become more critical if the image of the source document has unwanted characteristics, such as imperfections, backgrounds, printed text, shape, and variations in size. Another condition that can affect the quality of the attributes of handwritten signatures occurs when the image presents some distortion, such as perspective, inclination, scale, or unexpected resolution, all of them when scanning photos. All these interference can also harm the verification systems of handwritten signatures with the increase of false positives or false negatives in the classification process.

In this work, we propose an approach to the pixel-level segmentation of handwritten signatures on images. Our model has been trained with ID document images with the same characteristics and interference that can arise in a real-world scenario. With this, our model will be able to get around the problems presented during the capture of signature images in different identification documents in noisy environments. Our proposal will also enable the acquisition of signatures with greater fidelity in the strokes regardless of the types of pen, ink, background, preserving the graphic characteristics. Another contribution is that the preservation of the characteristics of the signature features will also make it possible to carry out graphotechnical analyzes. These features are used by forensic experts and may be applied in future systems for verifying handwritten signatures with a bias in forensic science.

We use a Fully Convolutional Network (FCN)\cite{long2015fully} for signature segmentation on identity document images with refinement layers for the alpha channel of the image.

The remainder of this paper is organized as follows: The Section \ref{related_works} presents the Related Works; in Section \ref{proposed_method} we describe the Proposed system; Section \ref{materials_methods} presents the Material and Methods; in Section \ref{results} we present the results and analysis of our proposal; and finally, the Section \ref{conclusions} depict the conclusions obtained from this work.

\section{Related Works} \label{related_works}

Handwritten signatures can be analyzed by online verification systems when an analysis is carried out during its production. In the most recent work by \cite{2020}, an online handwritten signature verification system based on a critical segment is proposed. The system identifies and exploits the segments that remain unchanged in the signatures to capture the intrinsic behavior in the signature incorporated in the signatures of each signatory. Another way to perform the signature analysis is offline when the signature has already been produced by the signatory. In \cite{2019}, a model based on Convolutional Neural Networks inspired by the architecture of Inception V1 is presented to learn about the characteristics existing in genuine signatures and forged signatures. The study uses offline subscriptions to public databases such as CEDAR. Our work is focused on offline subscriptions.

Studies have been carried out to investigate the performance of the Deep Learning algorithms from literature facing the task of signature and logo detection. The deep learning-based object detectors, namely, Faster R-CNN,  \textit{ZF}, \textit{VGG16}, $VGG_M$ and YOLOv2 where examined for this task. The proposed approach detects Signatures and Logos simultaneously \cite{sharma2018signature}. Mainly, in that study, the authors worked to detect signatures rather than segmentation of signature traits.

Thus, bounding boxes were generated around the detected signatures and logos. The dataset used was the Tobacco-800\cite{CDIP}, which has a clear background and is composed of scanned documents comprising printed text, signatures, and logos.

Other scientific papers have presented methods for the stroke-based extraction of signatures from document images. The proposed approach in \cite{melo2018fully} is based on an FCN trained to learn, map, and extract the handwritten signatures from documents. Although the proposal achieves good results, the network architecture requires a fixed size (512 x 512) of the input images \cite{melo2018fully}.

In \cite{plm2019}, a similar approach is used for signature extraction in identification documents. For this, the authors used an optimized U-net network with less trainable parameters and input nodes than \cite{melo2018fully}. To increase the generalization of the model, the authors applied the data augmentation technique in the database, generating greater image diversity during training. The model proposed in \cite{plm2019} achieved higher rates than \cite{melo2018fully}, despite having fewer parameters. 

To compose the structure of the first stage of our FCN+RL model, we selected an approach similar to the one proposed by \cite{plm2019}. The proposal of \cite{plm2019} presents a segmentation model at the stroke level, with promising results for the objectives of the first stage of our system proposed in this work.

In fact, stroke pixel integrity is of great importance to the offline signature verification process. Maintaining this integrity to the maximum can increase confidence in Handwritten Signature Verification (HSV) systems, especially if more technical approaches, such as graphoscopy in forensic science, are used.

In this sense, proposed methods have been presented for feature extraction using Deep Convolutional Neural Network combined with the SVM classifier to writer-independent (WI) handwritten signature verification systems. The proposed approach described in \cite{souza2018writer} outperformed other WI-HSV methods from the literature and outperformed writer-dependent methods from literature in some Brazilian dataset. Nevertheless, both works, which reach the state-of-the-art in the HSV task, assume the image signature pixels are available in a clean area.

In this paper, we are proposing a robust handwritten signature segmentation method that can be used to detect and extract only the signature pixels in some document. As much more the signature pixels can be extracted without noisy and distortion, better will be the results achieved by any HSV system in the signature verification task.

\section{The FCN+RL proposed architecture} \label{proposed_method}

The handwritten signature segmentation task is performed by an approach using an FCN encoder-decoder network architecture along with the refinement layers (RL) for the alpha channel of the signature image. The FCN is based on the signature segmentation neural network architecture proposed by \cite{plm2019}, that uses the FCN U-net architecture \cite{b3}. This FCN U-net model is then improved by the addition of the RL model, motivated by the results obtained by Xu et al. \cite{xu2017deep}. 

\begin{figure*}[htb!]
    \centering
    \includegraphics[width=\linewidth]{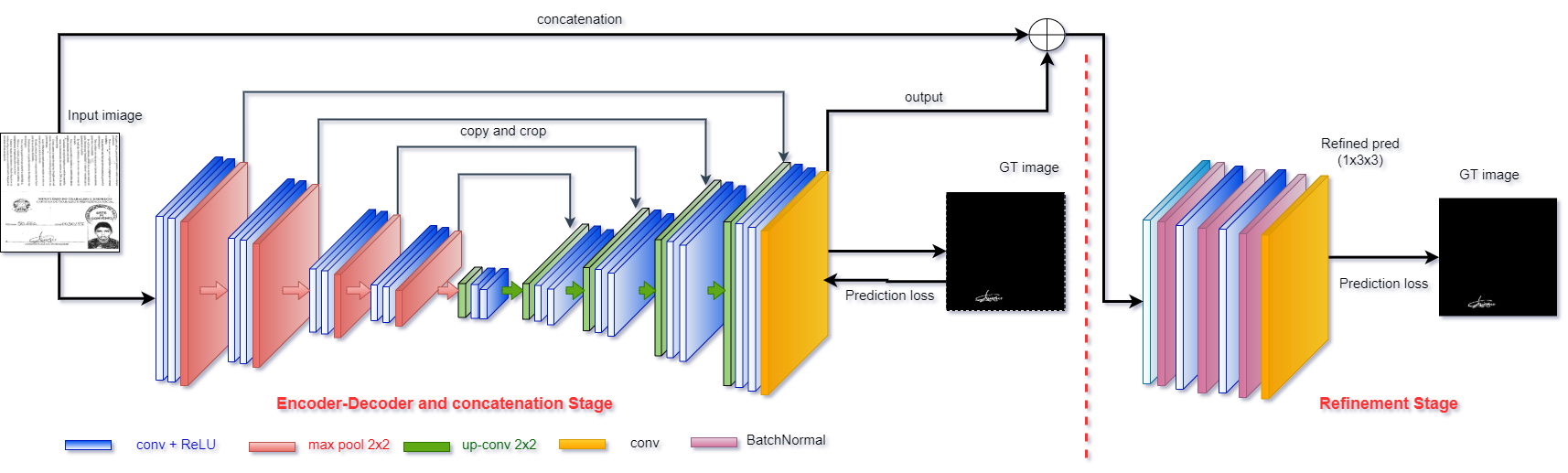}
    \caption{The FCN+RL proposed architecture. In this example, FCN performs segmentation by pixel classification, providing an image with the estimated signature pixels in the foreground. After the first stage, the concatenation between the input image and the image predicted by the FCN encoder-decoder layers is sent for input from the layer block for RL at stage two.}
    \label{fig:fcn_model}
\end{figure*}

Fig. \ref{fig:fcn_model} shows the architecture of our proposed model.

\subsection{The FCN Encoder-Decoder stage}

The convolutional operations (in blue) are set to $3\times3$ of size with ReLU activation function. The max-pooling layers (in red) are set to $2\times2$ with stride $2$. In the expansive path, there are upstream operations (in green) of size $2\times2$ concatenating with the corresponding characteristics of the path of contraction (gray arrows), followed by two convolutions of size $3\times3$ followed by ReLU operation. 

The training is carried out by two stages. First, we train the FCN encoder-decoder layers to learn the signature pixels.
At the FCN stage, the $512\times512$ image is sent to the network input that follows the contraction layers. Feature maps generated in the contraction layers are cropped and copy applied to concatenate the expansion layers. The expected output result is a binary image containing only the signature pixels predicted by the model. 

To calculate the error and adjust the weights, a ground-truth image is used. In this way, the model returns the error between the prediction and the ground truth images to fit the weights later. 

In  \cite{xu2017deep}, the authors use the original image and a corresponding trimap of the original images that are concatenated during the train process. These images then proceed to the first convolution layer. The disadvantage of using the trimap in our case is that the trimap image is also needed in the inference process. Therefore, in our model, we ignore the trimap and send only the original image to the input layer. Concatenation is performed between the contraction and expansion layers when  training the FCN. Thus, in the inference process, we use only the original image, to keep as much signature information as possible, such as the forensic experts would use in a real-world scenario.

\subsection{The Refinement Layer stage}

The RL model architecture consists of 4 convolutional layers. A non-linear ReLu layer follows each of the three first layers. Each convolution layer has a $64\times3\times3$ setting. Convolutional layers for refinement do not have max-pooling layers or upstream layers, so we add a Batch-normalization layer after each of the first three convolutional layers.

Before training the refinement layers, the weights of the FCN layers need to be frozen. The RL input image is generated using the concatenation between the original image and the output image predicted in the previous stage. This concatenation extends the alpha channel information and assists in the refinement process for the subscription region. 
The weight adjustment of RL block is performed using the same ground-truth image and procedure used for training the FCN block.

The  expected  result of the whole process is a binary image with the pixels of handwritten signatures as one and the any other irrelevant pixel as zero. Therefore, this binary image serves as a mask to select the pixels of the handwritten signatures in RGB or gray level from the original input image. 
\section{Materials and Methods}
\label{materials_methods}

\subsection{Datasets} \label{dataset}

Public datasets of handwritten signatures, such as MCYT and GPDS, do not meet the purpose of this work due to their composition because they do not show the poor image conditions that might impair the classification process. Another challenge for the acquisition of a database is the confidentiality of information, as these are documents with personal information. To overcome this drawbacks, the DSSigDataset-2 database \cite{plm2019} was used for the experiments in this work.

The DSSigDataset-2 is made up of $20,000$ document´s images with 200 background samples, and different distortion in the image. The handwritten signatures blended in the document image were selected from the MCYT dataset \cite{fierrez2004off} together with voluntarily generated signatures. 

\begin{figure}[!h]
  \centering
  \includegraphics[scale=0.15]{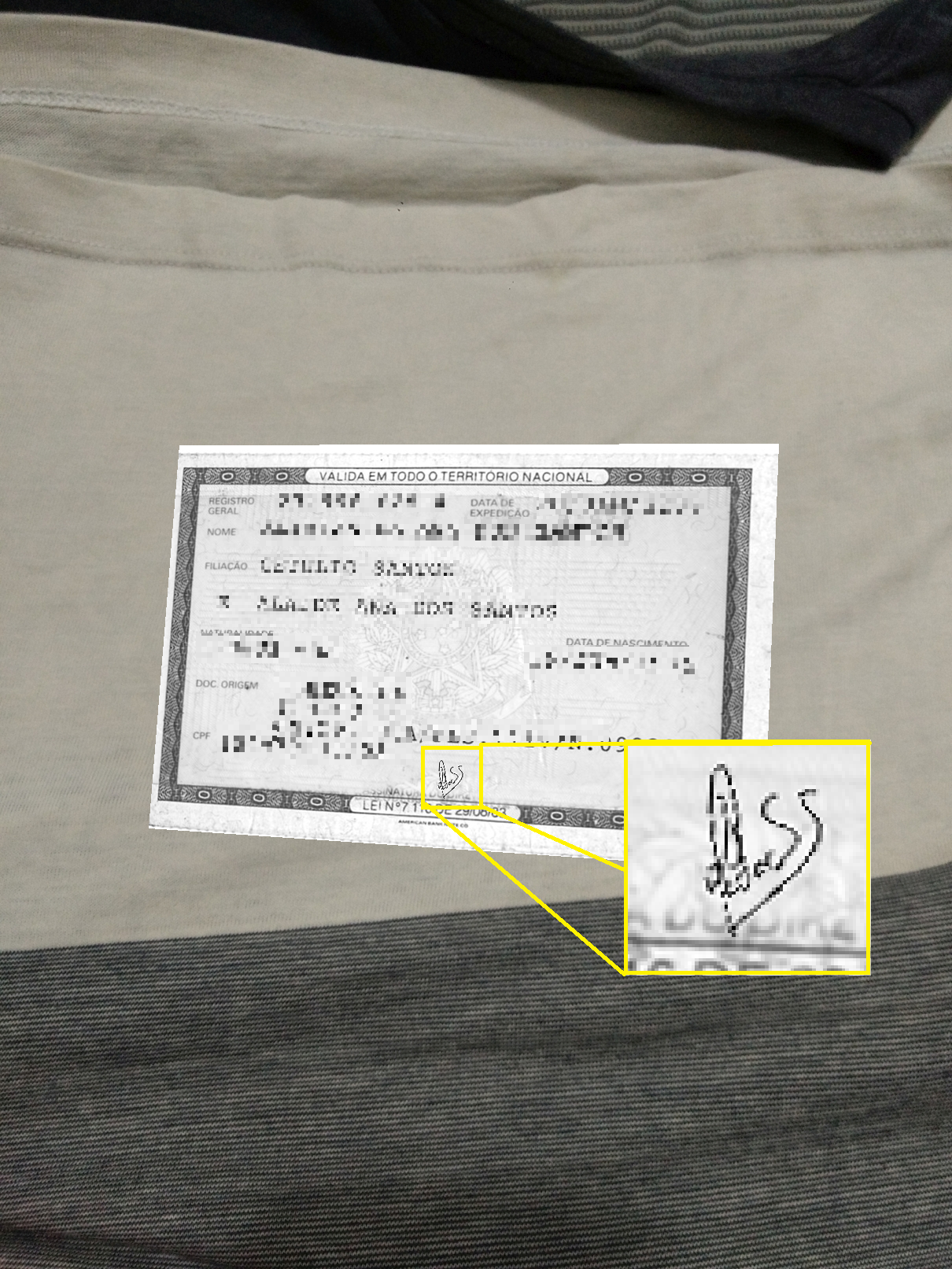}
  \caption{Example of an image of an identification document in the composition of the training database.}
  \label{fig:signature_small}
\end{figure}

Another aspect of the DSSigDataset-2 is the type of pen used, in which different types and different colors were used to avoid possible bias in the learning of the network regarding the color and type of ink. Fig. \ref{fig:signature_small} presents an example of an image from the DSSigDataset-2 database. A small area of the target handwritten signature is highlighted to give an idea of the challenge to detect the signature pixels in such conditions.

\subsection{Training procedure} \label{Train}

We split the 20,000 images of DSSigDataset-2 for training, validation, and test using the cross-validation method. We assigned $80\%$ for training and $15\%$ for validation, and $5\%$ for test, which respectively resulted in $16,000$ training images, $3,000$ validation images, and $1,000$ test images taking into account different handwritten signatures in all partitions. Also, since we had random image transformations applied in the document and the background, during the DSSigDataset-2 construction, we assure a complete unbiased dataset.

We used the Adam optimizer \cite{kingma2014adam} to minimize the objective function, which was the Dice coefficient (DC) \cite{b16}, shown in Equation \ref{eqDice}.

\begin{equation}
    DC = 2 \dfrac{|A \cap B|}{|A|+|B|}
    \label{eqDice}
\end{equation} where $A$ represents the ground-truth image and $B$ represents the segmented image at the network output.

\begin{figure}[h!]
    \centering
    \includegraphics[width=\linewidth]{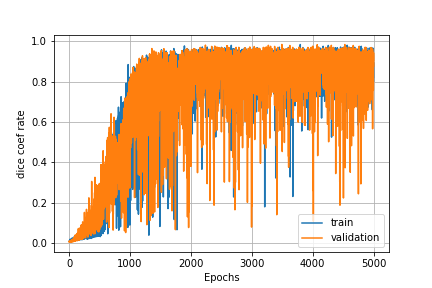}
    \caption{Graph of the evolution of the similarity rate for the evaluation metric with the Dice coefficient in the validation and training sets.}
    \label{fig:rateDice}
\end{figure}

\begin{figure}[h!]
    \centering
    \includegraphics[width=\linewidth]{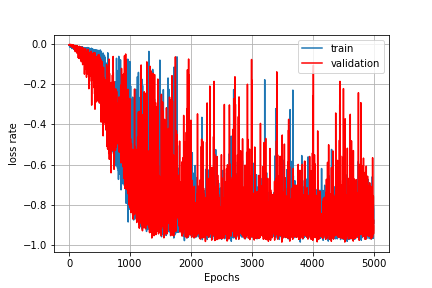}
    \caption{Graph of the evolution of the rate for objective function with the Dice coefficient in the validation and training sets.}
    \label{fig:lossDice}
\end{figure}

For training the FCN encoder-decoder (first stage), 10,000 epochs were used. For training the refinement layers, 5,000 epochs were used. Despite this number, our model has already obtained results with rates above 0.80 from epoch 1,000 for the data similarity rate between the output image and the ground-truth. Fig.\ref{fig:rateDice}  shows the similarity rate between the predicted and the expected data (pixels) for the Dice similarity coefficient (evaluation metric) during the training.
Parallel to the increase in the similarity rate, the model showed a loss rate consistent with the results. Fig.\ref{fig:lossDice} shows the evolution of the rate for the loss function (objective function).

\section{Results and Discussion} \label{results}

We performed several experiments to determine the best possible configuration and to validate the model's ability. We tested different configurations by evaluating the effects on optimization of hyper-parameters. However, we report in this paper the best validated configurations after all the preliminary experiments. 

To compare the results obtained from the experiments of this work, we implemented the model described in \cite{plm2019} as a baseline system. The reference model was also subjected to training with the DSSigDataset-2 database under the same conditions for the number of epochs and division of training and validation data. Tests applied to both models took place under the same conditions on the test set. 
Three metrics were used on the predictions of the models covered in this work: Structural Similarity (SSIM) index \cite{wang2004image}, Scale Invariant Feature Transform (SIFT) \cite{lowe2004distinctive}, and Dice Similarity Coefficient (DSC).
Quantitative results for similarity metrics are presented in Table \ref{tab:quant_result}. 


\begin{table}[h]
	\centering
	\caption{results for similarity metrics}
	\begin{tabular}{|l|c|c|c|c|c|c|}
		\hline
		\multicolumn{1}{|c|}{\multirow{2}{*}{Model}} & \multicolumn{2}{c|}{SSIM} & \multicolumn{2}{c|}{SIFT} & \multicolumn{2}{c|}{DSC} \\ \cline{2-7} 
		\multicolumn{1}{|c|}{} & rate & std & rate & std & rate & std \\ \hline
		FCN+RL & \textbf{96\%} & 0.04 & \textbf{97\%} & 0.05 & \textbf{88\%} & 0.18 \\ \hline
		Reference model\cite{plm2019} & 92\% & 0.06 & 56\% & 0.4 & 45\% & 0.39 \\ \hline
	\end{tabular}
    \label{tab:quant_result}
\end{table}

The FCN+RL model presents superior results for the three evaluation metrics. The results show different rates for different parameters. This observation is relevant to the characteristics considered by each metric used. The DSC technique presents lower values because it evaluates pixel-by-pixel of the entire region of the image and not just the morphology of the segmented area. Moreover, our proposed model is much more robust against scale and other distortions in the signature image. The RL block is responsible to filter out irrelevant pixels and filter in some pixels which cannot be detected by the FCN autoencoder block.

After performing the tests and applying the similarity metrics, three sets of data were selected with thirty samples for each set. These data were collected from the results of the SSIM, SIFT, and DSC tests and subjected to statistical analyses. 

First, we applied the Shapiro Wilk test to verify the normality of the data. This test indicates that the data do not follow a normal distribution for the datasets, so we applied the Wilcoxon-Mann-Whitney test. Tables \ref{tab:result_shapiro} and \ref{tab:result_wilcoxon} shows the results of the normality test and Wilcoxon test, respectively. 

For the null hypothesis, we consider that the means are the same for both models, and as an alternative hypothesis, we consider the difference in means between the two models. We considered a significance level of 0.05.

\begin{table}[h]
\centering
\caption{Normality test results}
\begin{tabular}{llllll}
\hline
 & \multicolumn{5}{c}{Shapiro Wilk test} \\ \hline
\multicolumn{1}{c}{} & \multicolumn{2}{c}{{ proposed model}} & \multicolumn{1}{c}{} & \multicolumn{2}{c}{{reference model\cite{plm2019}}} \\
\multicolumn{1}{c}{} & \multicolumn{1}{c}{W} & \multicolumn{1}{c}{p-value} & \multicolumn{1}{c}{} & \multicolumn{1}{c}{W} & \multicolumn{1}{c}{p-value} \\
SSIM & 0.659 & 4.25e-07 &  & 0.860 & 0.001 \\
SIFT & 0.510 & 6.93e-09 &  & 0.717 & 2.79e-06 \\
DSC & 0.600 & 7.32e-08 &  & 0.837 & 0.0003 \\ \hline
\end{tabular}
\label{tab:result_shapiro}
\end{table}

\begin{table}[h]
\centering
\caption{Results of the Wilcoxon-Mann-Whitney test.}
\begin{tabular}{lcl}
\hline
\multicolumn{3}{c}{Wilcoxon-MW} \\ \hline
\multicolumn{1}{c}{} & W & \multicolumn{1}{c}{p-value} \\
SSIM & 218.5 & 0.0003 \\
SIFT & 102.5 & 1.24e-7 \\
DSC & 158.5 & 8.03e-6 \\ \hline
\end{tabular}
\label{tab:result_wilcoxon}
\end{table}

The results of the Wilcoxon test show that the p-value values are lower than the level of significance. Therefore, we reject the null hypothesis. This result grants evidence that the models perform differently for the tests performed in this work. Observing the results of similarity rates between the two models, the statistical tests show evidence that our proposed model is statistically superior to the reference model.

Finally, we performed a qualitative assessment of the images predicted by the models. In this evaluation, it was possible to infer about the noise levels that negatively impact the subscription segmentation.

\begin{figure*}[!htb]
    \centering
    \begin{subfigure}[b]{0.26\textwidth}
        \centering
        \includegraphics[width=\textwidth]{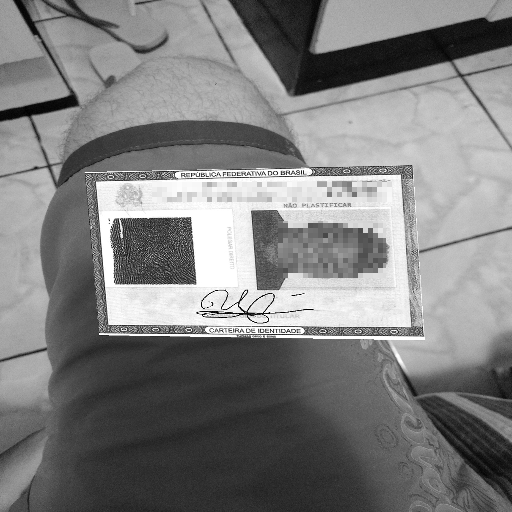}
        \caption{Input image signature.}
        \label{fig:signature_gt}
    \end{subfigure}
    \hfill
    \begin{subfigure}[b]{0.26\textwidth}
        \centering
        \includegraphics[width=\textwidth]{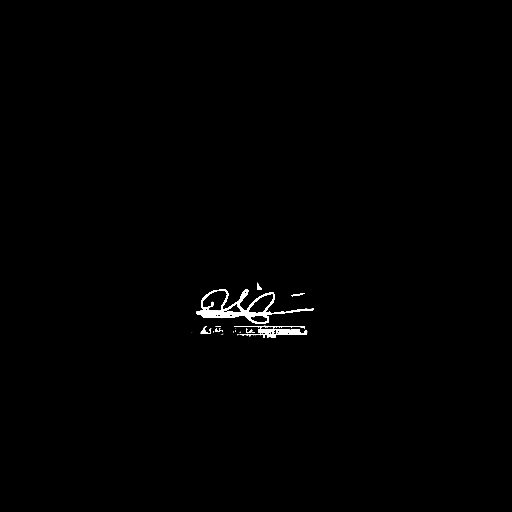}
        \caption{Reference model \cite{plm2019} result.}
        \label{fig:signature_gt}
    \end{subfigure}
    \hfill
    \begin{subfigure}[b]{0.26\textwidth}
        \centering
        \includegraphics[width=\textwidth]{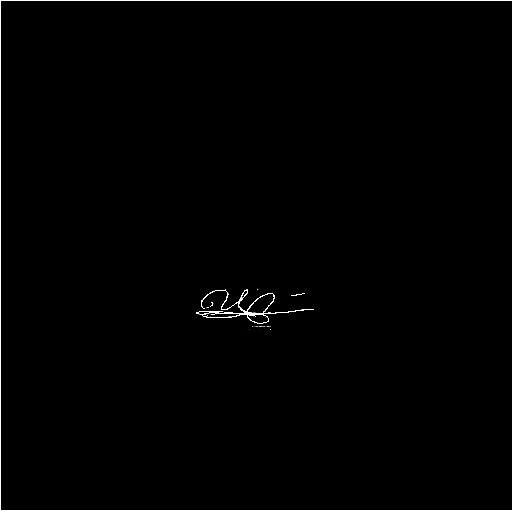}
        \caption{Our FCN+RL model result.}
    \end{subfigure}
    \hfill
    \begin{subfigure}[b]{0.26\textwidth}
        \centering
        \includegraphics[width=\textwidth]{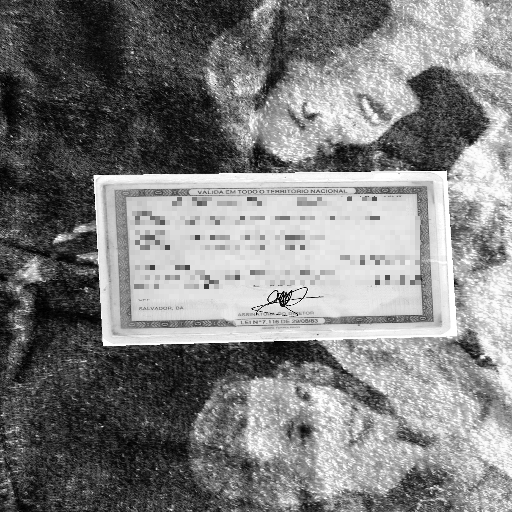}
        \caption{Input image signature.}
    \end{subfigure}
    \hfill
    \begin{subfigure}[b]{0.26\textwidth}
        \centering
        \includegraphics[width=\textwidth]{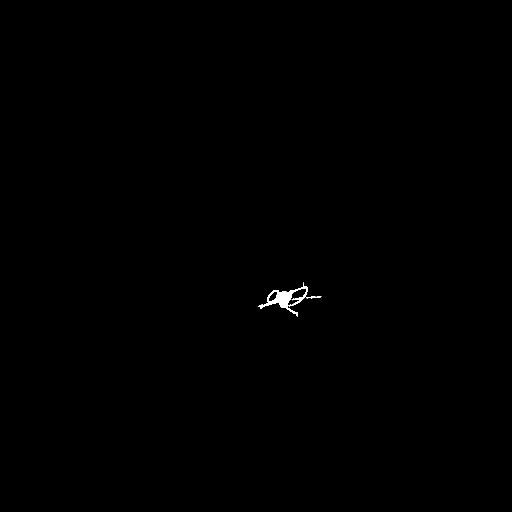}
        \caption{Reference model \cite{plm2019} result.}
    \end{subfigure}
    \hfill
    \begin{subfigure}[b]{0.26\textwidth}
        \centering
        \includegraphics[width=\textwidth]{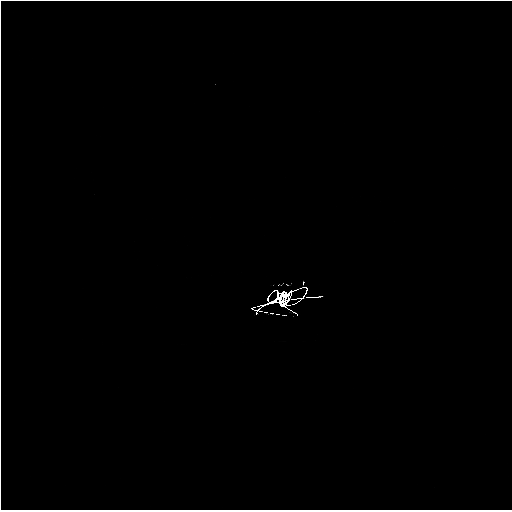}
        \caption{Our FCN+RL model result.}
    \end{subfigure}
    \hfill
    \caption{Qualitative result of the model used in \cite{plm2019} and our FCN+RL proposed model.}
    \label{fig:quali_result}
\end{figure*}

\begin{figure*}[!htb]
    \centering
    \begin{subfigure}[b]{0.26\textwidth}
        \centering
        \includegraphics[width=\textwidth]{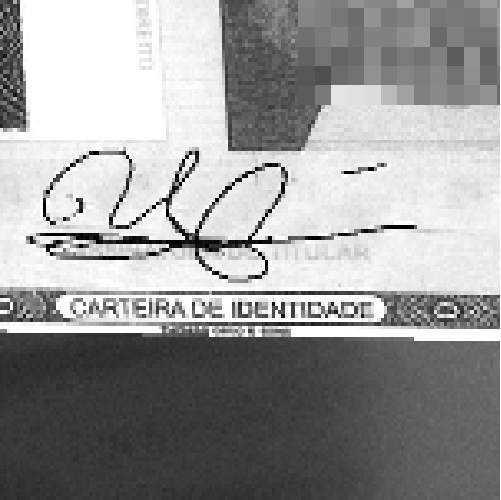}
        \caption{Input image signature.}
        \label{fig:signature_in}
    \end{subfigure}
    \hfill
    \begin{subfigure}[b]{0.26\textwidth}
        \centering
        \includegraphics[width=\textwidth]{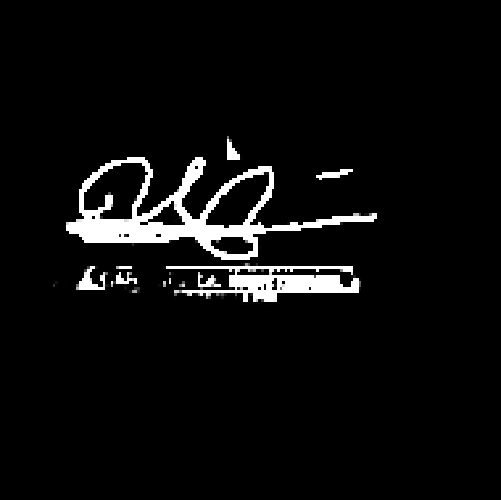}
        \caption{Reference model \cite{plm2019} result.}
        \label{fig:signature_gt}
    \end{subfigure}
    \hfill
    \begin{subfigure}[b]{0.26\textwidth}
        \centering
        \includegraphics[width=\textwidth]{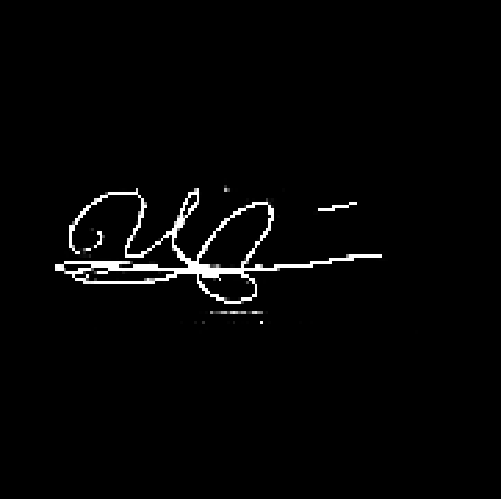}
        \caption{Our FCN+RL model result.}
    \end{subfigure}
    \hfill
    \begin{subfigure}[b]{0.26\textwidth}
        \centering
        \includegraphics[width=\textwidth]{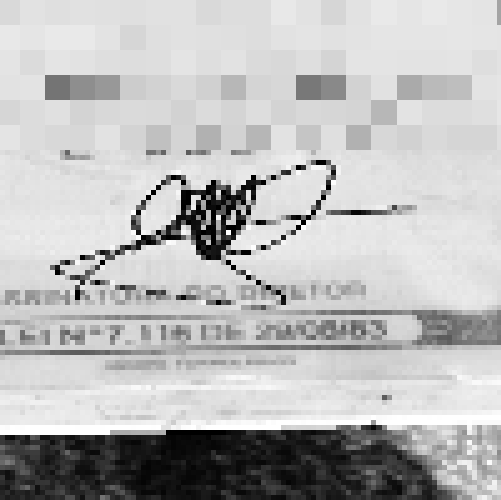}
        \caption{Input image signature.}
    \end{subfigure}
    \hfill
    \begin{subfigure}[b]{0.26\textwidth}
        \centering
        \includegraphics[width=\textwidth]{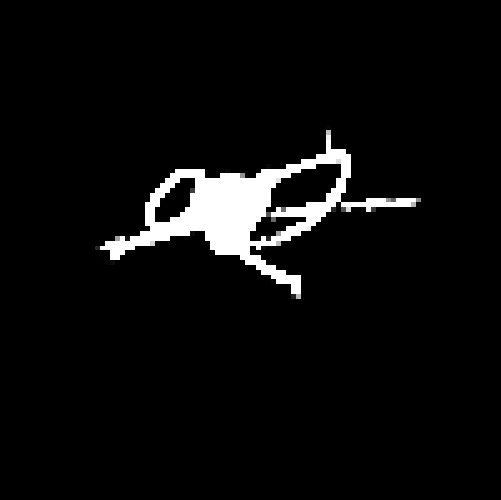}
        \caption{Reference model \cite{plm2019} result.}
    \end{subfigure}
    \hfill
    \begin{subfigure}[b]{0.26\textwidth}
        \centering
        \includegraphics[width=\textwidth]{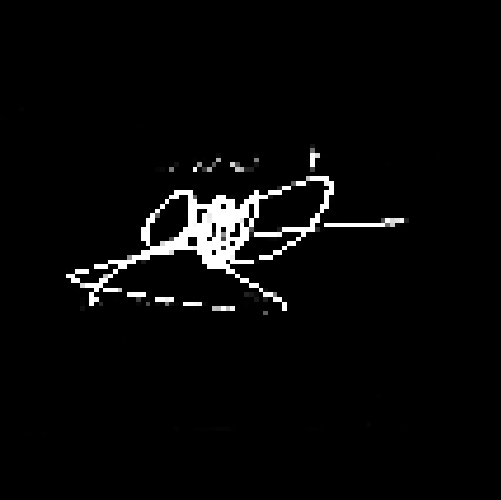}
        \caption{Our FCN+RL model result.}
    \end{subfigure}
    \hfill
    \caption{Zoom in the images shown in Fig. \ref{fig:quali_result}.}
    \label{fig:zoom_quali_result}
\end{figure*}

Qualitative results show an improvement in the segmentation of handwritten signatures by our FCN+RL model. Fig. \ref{fig:quali_result} shows the results for the two models, the reference model \cite{plm2019} and our model FCN+RL. Fig. \ref{fig:zoom_quali_result} shows the enlarged results of the subscriptions. It is easy to observe the results produced by our FCN+RL model show greater fidelity to the original image, in addition to having less background noise and keeping the fine details of the signatures in both examples.

\section{Conclusions} \label{conclusions}

Extracting handwritten signatures from images with a complex background and noise interference, such as identification documents, is a complex but promising task for the application of signature verification systems. Achieving the maximum fidelity of a signature's characteristics can have a positive impact on the classification results, including guaranteeing the graphotechnical characteristics used by forensic specialists.

In this article, we proposed an approach to an FCN encoder-decoder network with refinement layers using a concatenation of the alpha channel of the region of interest on the original image for the segmentation of handwritten signatures, the FCN+RL. 
The technique used with the alpha channel opacity over the image in the second stage of the convolution layers, refinement stage, provided an increase in the similarity rate between the predicted images and the ground truth. This refinement reduces the scattering of the region of interest and presents the pixels of the signatures much less sparse and greater preservation of the graphic characteristics of the signatures. 

With this result, signature verification systems can be used on handwritten signatures extracted from different images of ID documents captured by various computing devices such as smartphones. 
Catch signatures will be much cleaner and preserved in the segmentation process. In this way, more technical analyzes, such as those used by forensic science, may make use of the graphotechnical characteristics held in the signatures, even submitted in digital systems. 

The evaluation metrics used were SSIM, SIFT, and the Data Similarity Coefficient. To compare the results obtained with the proposed approach, we replicate the subscription segmentation model proposed by \cite{plm2019} to use as a reference. 

The best results of our proposed model obtained an improvement rate of more than 40 percentage points about the reference model. The qualitative results show a greater fidelity in the characteristics of the signature and a low level of background noise.

In the next works, we intend to present the signature verification models that are being developed by us and will be used for the calculation of graphotechnical analyzes.

As possible future works, it is also possible to point out the implementation of a semantic segmentation solution based on the proposed architecture for the application domain relevant to the structure and characteristics of documents. An example is the detection of anomalies in documents forged or created by gross forgeries—interest area of the forensic science, the documentoscopy. 

In addition to the implementation, as mentioned earlier, instantiating the model to other application domains would allow a way to assess the generalization capacity of the proposed model, as well as to detect possible adjustable points in the architecture.

\section*{Acknowledgment}
This study was financed in part by: Coordenação de Aperfeiçoamento de Pessoal de Nível Superior - Brasil (CAPES) - Finance Code 001, FACEPE, and CNPq - Brazilian research agencies.

\bibliographystyle{IEEEtran}

\bibliography{paper}

\begin{thebibliography}{10}
\providecommand{\url}[1]{#1}
\csname url@samestyle\endcsname
\providecommand{\newblock}{\relax}
\providecommand{\bibinfo}[2]{#2}
\providecommand{\BIBentrySTDinterwordspacing}{\spaceskip=0pt\relax}
\providecommand{\BIBentryALTinterwordstretchfactor}{4}
\providecommand{\BIBentryALTinterwordspacing}{\spaceskip=\fontdimen2\font plus
\BIBentryALTinterwordstretchfactor\fontdimen3\font minus
  \fontdimen4\font\relax}
\providecommand{\BIBforeignlanguage}[2]{{%
\expandafter\ifx\csname l@#1\endcsname\relax
\typeout{** WARNING: IEEEtran.bst: No hyphenation pattern has been}%
\typeout{** loaded for the language `#1'. Using the pattern for}%
\typeout{** the default language instead.}%
\else
\language=\csname l@#1\endcsname
\fi
#2}}
\providecommand{\BIBdecl}{\relax}
\BIBdecl

\bibitem{melo2018fully}
V.~K. S.~L. Melo and B.~B.~L. Dantas, ``A fully convolutional network for
  signature segmentation from document images,'' in \emph{2018 16th
  International Conference on Frontiers in Handwriting Recognition
  (ICFHR)}.\hskip 1em plus 0.5em minus 0.4em\relax IEEE, 2018, pp. 540--545.

\bibitem{pansare2012handwritten}
A.~Pansare and S.~Bhatia, ``Handwritten signature verification using neural
  network,'' \emph{International Journal of Applied Information Systems},
  vol.~1, no.~2, pp. 44--49, 2012.

\bibitem{coetzer2004offline}
J.~Coetzer, B.~M. Herbst, and J.~A. du~Preez, ``Offline signature verification
  using the discrete radon transform and a hidden markov model,'' \emph{EURASIP
  Journal on applied signal processing}, vol. 2004, pp. 559--571, 2004.

\bibitem{matsuda2016effective}
K.~Matsuda, W.~Ohyama, T.~Wakabayashi, and F.~Kimura, ``Effective
  random-impostor training for combined segmentation signature verification,''
  in \emph{2016 15th International Conference on Frontiers in Handwriting
  Recognition (ICFHR)}.\hskip 1em plus 0.5em minus 0.4em\relax IEEE, 2016, pp.
  489--494.

\bibitem{anikin2016handwritten}
I.~V. Anikin and E.~S. Anisimova, ``Handwritten signature recognition method
  based on fuzzy logic,'' in \emph{2016 Dynamics of Systems, Mechanisms and
  Machines (Dynamics)}.\hskip 1em plus 0.5em minus 0.4em\relax IEEE, 2016, pp.
  1--5.

\bibitem{ren2015faster}
S.~Ren, K.~He, R.~Girshick, and J.~Sun, ``Faster r-cnn: Towards real-time
  object detection with region proposal networks,'' in \emph{Advances in neural
  information processing systems}, 2015, pp. 91--99.

\bibitem{redmon2017yolo9000}
J.~Redmon and A.~Farhadi, ``Yolo9000: better, faster, stronger,'' in
  \emph{Proceedings of the IEEE conference on computer vision and pattern
  recognition}, 2017, pp. 7263--7271.

\bibitem{sharma2018signature}
N.~Sharma, R.~Mandal, R.~Sharma, U.~Pal, and M.~Blumenstein, ``Signature and
  logo detection using deep cnn for document image retrieval,'' in \emph{2018
  16th International Conference on Frontiers in Handwriting Recognition
  (ICFHR)}.\hskip 1em plus 0.5em minus 0.4em\relax IEEE, 2018, pp. 416--422.

\bibitem{guerbai2015effective}
Y.~Guerbai, Y.~Chibani, and B.~Hadjadji, ``The effective use of the one-class
  svm classifier for handwritten signature verification based on
  writer-independent parameters,'' \emph{Pattern Recognition}, vol.~48, no.~1,
  pp. 103--113, 2015.

\bibitem{hafemann2017learning}
L.~G. Hafemann, R.~Sabourin, and L.~S. Oliveira, ``Learning features for
  offline handwritten signature verification using deep convolutional neural
  networks,'' \emph{Pattern Recognition}, vol.~70, pp. 163--176, 2017.

\bibitem{soleimani2016deep}
A.~Soleimani, B.~N. Araabi, and K.~Fouladi, ``Deep multitask metric learning
  for offline signature verification,'' \emph{Pattern Recognition Letters},
  vol.~80, pp. 84--90, 2016.

\bibitem{diaz2016generation}
M.~Diaz, M.~A. Ferrer, G.~S. Eskander, and R.~Sabourin, ``Generation of
  duplicated off-line signature images for verification systems,'' \emph{IEEE
  transactions on pattern analysis and machine intelligence}, vol.~39, no.~5,
  pp. 951--964, 2016.

\bibitem{vargas2007off}
F.~Vargas, M.~Ferrer, C.~Travieso, and J.~Alonso, ``Off-line handwritten
  signature gpds-960 corpus,'' in \emph{Ninth International Conference on
  Document Analysis and Recognition (ICDAR 2007)}, vol.~2.\hskip 1em plus 0.5em
  minus 0.4em\relax IEEE, 2007, pp. 764--768.

\bibitem{ortega2003mcyt}
J.~Ortega-Garcia, J.~Fierrez-Aguilar, D.~Simon, J.~Gonzalez, M.~Faundez-Zanuy,
  V.~Espinosa, A.~Satue, I.~Hernaez, J.-J. Igarza, C.~Vivaracho \emph{et~al.},
  ``Mcyt baseline corpus: a bimodal biometric database,'' \emph{IEE
  Proceedings-Vision, Image and Signal Processing}, vol. 150, no.~6, pp.
  395--401, 2003.

\bibitem{diaz2019perspective}
M.~Diaz, M.~A. Ferrer, D.~Impedovo, M.~I. Malik, G.~Pirlo, and R.~Plamondon,
  ``A perspective analysis of handwritten signature technology,'' \emph{ACM
  Computing Surveys (CSUR)}, vol.~51, no.~6, pp. 1--39, 2019.

\bibitem{mello2012}
C.~A.~B. MELLO, W.~P. SANTOS, and A.~L.~I. OLIVEIRA, \emph{Digital Document
  Analysis and Processing}.\hskip 1em plus 0.5em minus 0.4em\relax New York:
  Nova Science Publishers Inc, 2012.

\bibitem{long2015fully}
J.~Long, E.~Shelhamer, and T.~Darrell, ``Fully convolutional networks for
  semantic segmentation,'' in \emph{Proceedings of the IEEE conference on
  computer vision and pattern recognition}, 2015, pp. 3431--3440.

\bibitem{2020}
Y.~{Ren}, C.~{Wang}, Y.~{Chen}, M.~C. {Chuah}, and J.~{Yang}, ``Signature
  verification using critical segments for securing mobile transactions,''
  \emph{IEEE Transactions on Mobile Computing}, vol.~19, no.~3, pp. 724--739,
  2020.

\bibitem{2019}
R.~K. {Mohapatra}, K.~{Shaswat}, and S.~{Kedia}, ``Offline handwritten
  signature verification using cnn inspired by inception v1 architecture,'' in
  \emph{2019 Fifth International Conference on Image Information Processing
  (ICIIP)}, 2019, pp. 263--267.

\bibitem{CDIP}
\BIBentryALTinterwordspacing
G.~Agam, S.~Argamon, O.~Frieder, D.~Grossman, and D.~Lewis, \emph{The Complex
  Document Image Processing (\uppercase{CDIP}) test collection}, Illinois
  Institute of Technology, 2006. [Online]. Available:
  \url{http://ir.iit.edu/projects/CDIP.html}
\BIBentrySTDinterwordspacing

\bibitem{plm2019}
P.~{G. S. Silva}, C.~{A. M. Lopes Junior}, E.~{Lima}, B.~{Leite Dantas
  Bezerra}, and C.~{Zanchettin}, ``Speeding-up the handwritten signature
  segmentation process through an optimized fully convolutional neural
  network,'' in \emph{2019 15th International Conference on Document Analysis
  and Recognition (ICDAR)}, Sep 2019.

\bibitem{souza2018writer}
V.~L. Souza, A.~L. Oliveira, and R.~Sabourin, ``A writer-independent approach
  for offline signature verification using deep convolutional neural networks
  features,'' in \emph{2018 7th Brazilian Conference on Intelligent Systems
  (BRACIS)}.\hskip 1em plus 0.5em minus 0.4em\relax IEEE, 2018, pp. 212--217.

\bibitem{b3}
O.~Ronneberger, P.~Fischer, and T.~Brox, ``U-net: Convolutional networks for
  biomedical image segmentation,'' in \emph{International Conference on Medical
  image computing and computer-assisted intervention}.\hskip 1em plus 0.5em
  minus 0.4em\relax Springer, 2015, pp. 234--241.

\bibitem{xu2017deep}
N.~Xu, B.~Price, S.~Cohen, and T.~Huang, ``Deep image matting,'' in
  \emph{Proceedings of the IEEE Conference on Computer Vision and Pattern
  Recognition}, 2017, pp. 2970--2979.

\bibitem{fierrez2004off}
``Um sistema de verificação de assinaturas off-line baseado na fusão de
  informações locais e globais,'' in \emph{Workshop Internacional de
  Autenticação Biométrica}.

\bibitem{kingma2014adam}
D.~P. Kingma and J.~Ba, ``Adam: A method for stochastic optimization,''
  \emph{arXiv preprint arXiv:1412.6980}, 2014.

\bibitem{b16}
A.~A. Novikov, D.~Major, D.~Lenis, J.~Hladuvka, M.~Wimmer, and K.~Bühler,
  ``Fully convolutional architectures for multi-class segmentation in chest
  radiographs.'' \emph{CoRR}, vol. abs/1701.08816, 2017.

\bibitem{wang2004image}
Z.~Wang, A.~C. Bovik, H.~R. Sheikh, and E.~P. Simoncelli, ``Image quality
  assessment: from error visibility to structural similarity,'' \emph{IEEE
  transactions on image processing}, vol.~13, no.~4, pp. 600--612, 2004.

\bibitem{lowe2004distinctive}
D.~G. Lowe, ``Distinctive image features from scale-invariant keypoints,''
  \emph{International journal of computer vision}, vol.~60, no.~2, pp. 91--110,
  2004.

\end{thebibliography}


\end{document}